\documentclass[letterpaper]{article} % DO NOT CHANGE THIS
\usepackage{aaai24}  % DO NOT CHANGE THIS

\usepackage{times}  % DO NOT CHANGE THIS
\usepackage{helvet}  % DO NOT CHANGE THIS
\usepackage{courier}  % DO NOT CHANGE THIS
\usepackage[hyphens]{url}  % DO NOT CHANGE THIS
\usepackage{graphicx} % DO NOT CHANGE THIS
\usepackage{natbib}  % DO NOT CHANGE THIS AND DO NOT ADD ANY OPTIONS TO IT
\usepackage{caption} % DO NOT CHANGE THIS AND DO NOT ADD ANY OPTIONS TO IT
\usepackage{amsmath}
\usepackage{amssymb}

\usepackage{subcaption} 
\usepackage{booktabs}
\usepackage{color}

\usepackage{algorithm}
\usepackage{algorithmic}

\usepackage{newfloat}
\usepackage{listings}
\DeclareCaptionStyle{ruled}{labelfont=normalfont,labelsep=colon,strut=off} % DO NOT CHANGE THIS
\floatstyle{ruled}
\newfloat{listing}{tb}{lst}{}
\floatname{listing}{Listing}
 \nocopyright %-- Your paper will not be published if you use this command
\title{A Unified Masked Autoencoder with Patchified Skeletons for Motion Synthesis}
\author{
    Esteve Valls Mascar\'o \textsuperscript{\rm 1},
    Hyemin Ahn \textsuperscript{\rm 2},
    Dongheui Lee\textsuperscript{\rm 1,\rm 3},
}
\affiliations{
    \textsuperscript{\rm 1} Technische Universität Wien (TUWien), 
    \textsuperscript{\rm 2} UNIST, 
    \textsuperscript{\rm 3} German Aerospace Center (DLR) \\
   \small{ \texttt{\{esteve.valls.mascaro, dongheui.lee\}@tuwien.ac.at}, \texttt{hyemin.ahn@unist.ac.kr}}
}

\usepackage{bibentry}
\begin{document}

\maketitle

\begin{abstract}
The synthesis of human motion has traditionally been addressed through task-dependent models that focus on specific challenges, such as predicting future motions or filling in intermediate poses conditioned on known key-poses. In this paper, we present a novel task-independent model called UNIMASK-M, which can effectively address these challenges using a unified architecture. Our model obtains comparable or better performance than the state-of-the-art in each field. Inspired by Vision Transformers (ViTs), our UNIMASK-M model decomposes a human pose into body parts to leverage the spatio-temporal relationships existing in human motion. Moreover, we reformulate various pose-conditioned motion synthesis tasks as a reconstruction problem with different masking patterns given as input. By explicitly informing our model about the masked joints, our UNIMASK-M becomes more robust to occlusions. Experimental results show that our model successfully forecasts human motion on the Human3.6M dataset. Moreover, it achieves state-of-the-art results in motion inbetweening on the LaFAN1 dataset, particularly in long transition periods. More information can be found on the project website \url{https://sites.google.com/view/estevevallsmascaro/publications/unimask-m} .
\end{abstract}

%%%%%%%%% BODY TEXT
\section{Introduction}

\begin{figure}
\centering
% \includegraphics[width=0.48\textwidth]{pics/overall_goal_transp_font.jpg}
% \caption{\textbf{Visual representation of the different human motion synthesis tasks.} The aim of this work is to design a model that handles human motion forecasting, inbetweening, and reconstruction, with its subvariations. Red and blue skeletons denote a known skeleton joint, while light gray indicates
% the joint is masked.}
\includegraphics[width=0.47\textwidth]{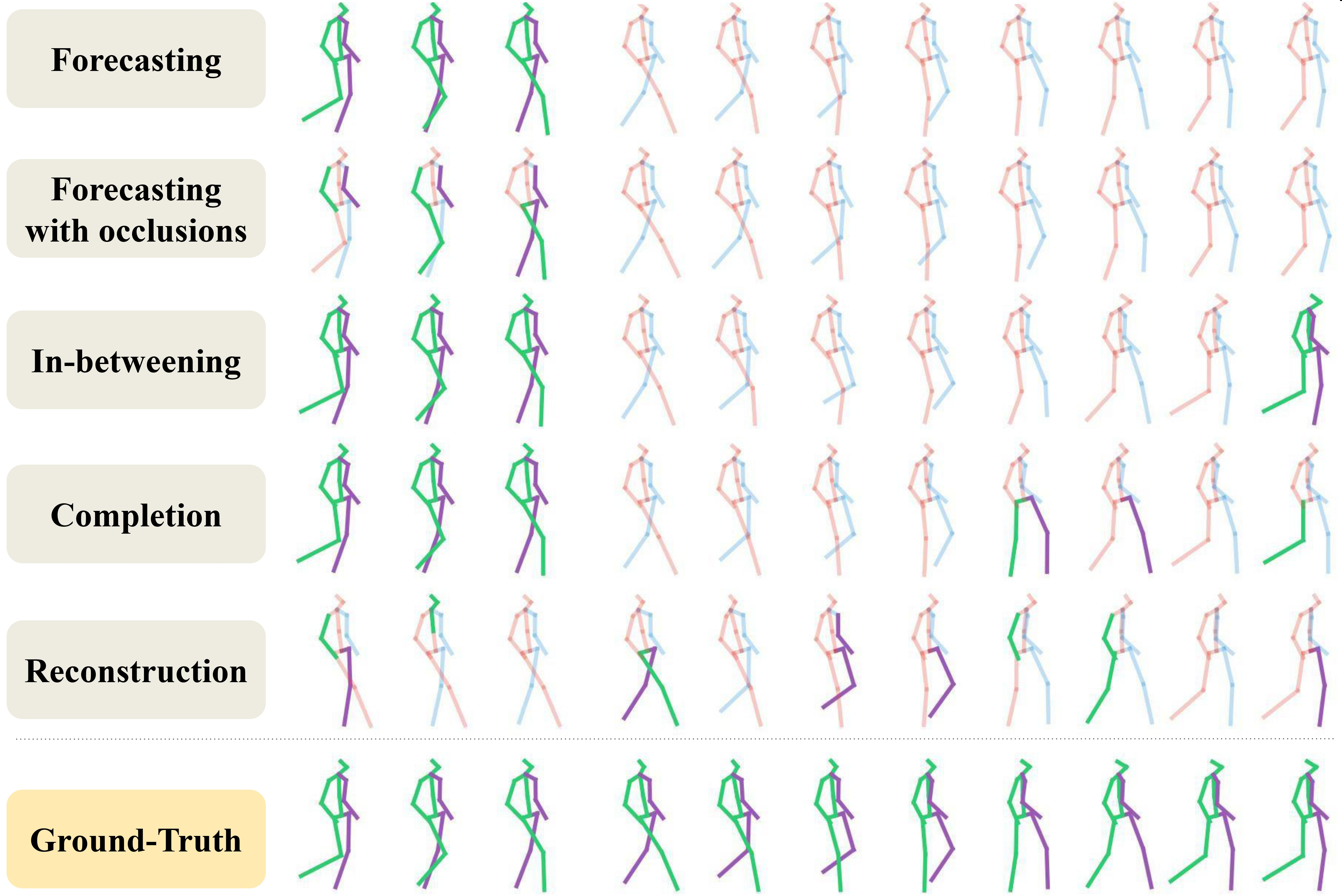}
\caption{\textbf{Unified architecture for different human motion synthesis tasks.} Green and purple skeletons denote a known skeleton joint, while light red and green represents our model prediction over a masked joint.}
\label{fig:overall_goal}
\end{figure}

Synthesizing plausible human motions is a long-standing challenge in the computer vision and graphics community. Due to the high complexity of the field, researchers often focus on addressing only a specific subset of human motion synthesis, as the different tasks involve varied input-output patterns for the model. Unlike previous works, we propose a unified architecture for solving various motion synthesis tasks. These different tasks are depicted in Fig. \ref{fig:overall_goal} and represent the motivation for the research addressed by this paper.

For instance, the 3D human motion forecasting task aims to predict the human's subsequent 3D body trajectory based only on past observations. In this field, researchers restrict their model to learning in a causality manner \cite   {sRNN, ST-TR, 2CH-TR}, where the new pose only depends on the previous frames. However, these causal-based approaches are not well-suited for the motion inbetweening task, which involves completing the body poses between the past observations and a known future pose. Most inbetweening models \cite{TRANSFORMER_MIB1,CMIB, DELTA_INTERP} usually leverage bidirectional temporal relationships since both past and future poses are key for filling the middle poses. However, these works consider the full-body skeleton as a unique condition, constraining the generation process. In our work, we additionally consider the motion completion task, where only independent body parts constrain the motion synthesis in the future, which enhances the freedom in the generation. Finally, motion reconstruction is the task of recovering a given motion subjected to occlusions, as might occur in real-world applications \cite{HRI_MOTION}. Dissecting the motion synthesis field in all these different tasks diminishes the versatility of the models. To overcome this challenge, generative approaches \cite{UnifiedCVAE, MDM} have been proposed recently to tackle motion synthesis with a unified stochastic model, but requires high computational resources and are unsuitable for real-time prediction \cite{MDM} or underperform compared to deterministic models \cite{UnifiedCVAE}, as also indicated by \cite{DiffPred:2023}.

In this work, we consider both efficiency and versatility as essential characteristics of a model, which should achieve high performance across multiple tasks. Inspired by the recent success of masked autoencoders in reconstructing images \cite   {MAE, beit}, and videos \cite   {videomae}, we adapt the motion synthesis as a reconstruction problem: to recover a sequence of masked human skeletons regardless of the masking pattern. In addition, we propose a deconstruction model to decompose human skeletons into body-part patches (i.e., legs, arms, trunk), which boosts the model's performance while giving more flexibility. Thanks to our deconstruction approach, our model can synthesize motion by leveraging the spatio-temporal relationships of partial body parts through self-attention \cite{transformer}. Therefore, our \textbf{UNI}fied \textbf{M}asked \textbf{A}utoencoder with patchified \textbf{SK}eletons for \textbf{M}otion synthesis, called \textbf{UNIMASK-M}, serves as the first model to address the deterministic motion synthesis task as a reconstruction problem through pose decomposition and masked autoencoders, and achieves comparable performance to the state-of-the-art in several motion synthesis tasks. The contributions of our paper can be summarized in three-fold: 

\begin{enumerate}
    \item A unique model that achieves performance comparable to the state-of-the-art in various motion synthesis tasks.
    \item A pose decomposition approach that deconstructs a single skeleton into patches, thus giving more freedom in the pose conditioning while boosting the performance.
    \item An efficient model suitable for real-time human motion synthesis and robust to occlusions.
\end{enumerate}

%-------------------------------------------------------------------------
\section{Related Work}
Literature has addressed each human motion synthesis task as a distinct challenge. This section first provides an overview of each motion synthesis task. Then, we describe the Masked Autoencoders which are the core of our model.

\subsection{Human Motion Synthesis}

Motion synthesis conditioned by key-poses serves as a basis for understanding observed humans' intentions or enabling agents or robots to collaborate with them.

\textbf{Motion Forecasting.} The aim is to predict future human motion based only on past observations. Early works \cite   {RNN_Motion2, sRNN} employed Recurrent Neural Networks (RNNs) to process the time-series dependencies of poses. Subsequently, \cite   {DCT-GCN, ST-DGCN} proposed Discrete Cosine Transformations (DCT) to encode temporal information and adopt Graph Convolutional Networks \cite   {gcn} to model the spatial relationships. Recent works \cite   {2CH-TR, mao2020history} adopt the Transformer-based model \cite   {transformer} to leverage the self-attention among joint information in both space and time. Recently, \cite   {SIMPLE} has introduced the use of temporal Multi-Layer Perceptrons (MLP) in an autoregressive process. Despite significant advances in performance and efficiency, all these models cannot deal with any motion synthesis task if future key-poses condition the motion.

\textbf{Motion Inbetweening.} In this task, the model is given a sequence of past poses and a future target pose and the goal is to fill in the motion between these key-poses. Early studies \cite   {RNN1_MIB} used RNNs to model the temporal relationships between the skeletons. \cite   {LSTM_MIB_TG} introduced scheduled noise injection in the target pose to encourage motion variations. \cite   {TRANSFORMER_MIB1} adopted a self-attention block to refine the interpolated poses and a key-pose embedding to support multiple whole pose conditions in a sequence. More recently, \cite   {DELTA_INTERP} adapted cross-attention to synthesize the missing frames. \cite   {TRANSFORMER_MIB1, DELTA_INTERP} referenced the predicted motion to an interpolated sequence between key-poses, thus reformulating the inbetweening problem as the synthesis of the variation of the poses from an interpolated reference. \cite   {CMIB} highlights the motivation of conditioning the motion synthesis through partial body poses but still requires the last pose to be full-body for sequence completion. Inspired by the image inpainting strategy using Denoising Diffusion Probabilistic Models (DDPMs) \cite   {ddpm_paper}, \cite   {MDM} evaluates their model for human motion completion conditioned by key-poses. However, their work requires high computation, and the denoising process impedes real-time execution. Unlike other works, we propose an efficient model that can deal with motion inbetweening and completion, independently of having the totality of the key-poses.

\textbf{Motion Reconstruction} In real scenarios, the estimated human poses can be subjected to occlusions. However, the aforementioned models overlook this problem by projecting the full-body skeletons into a single representation space. Therefore, the model cannot effectively handle partial occlusions within a single pose, just the complete absence of the full-body skeleton. \cite{dual_maskedAE, motionbert, posebert} adopt masked modeling for motion reconstruction in 3D human skeleton estimation. However, \cite{dual_maskedAE, motionbert} cannot handle partial pose masking, and \cite{posebert} uses joint-level masking, which we show that underperforms.

\textbf{Unified Motion Synthesis. } Embracing motion synthesis only accounting for particular masking patterns allows researchers to design models that can excel at one of the specific tasks. However, designing a unified architecture that performs well across multiple tasks of motion synthesis is challenging. To tackle this challenge, \cite   {UnifiedCVAE} uses a Conditional Variational Autoencoder (CVAE) \cite   {vae} to sample from the masked sequence distribution and condition the generated missing poses by the anchor poses. \cite   {MDM} evaluated their DDPM architecture in different motion synthesis tasks. However, we prioritize deterministic over stochastic approaches as we observe that diversity in motions can be obtained by adding noise to the anchor poses \cite   {LSTM_MIB_TG}, while having closer predictions to the reality is essential for real scenarios. 

\textbf{Semantic conditioned Motion Generation.} While previous tasks tackle motion synthesis conditioned by key-poses, other modalities can be used, such as text \cite   {MDM}, or 3D scenes \cite   {gimo}. Despite considering these works to be relevant, we focus on the unification of pose-conditioned human motion synthesis, which can serve as a basis to later adopt in other modalities.

\subsection{Masked Autoencoders}
Learning robust visual representations is essential for the image domain. With the success of Vision transformers (ViT) \cite   {vit}, masked visual modeling has emerged as a key design to encode images effectively.  \cite   {beit} proposed to learn visual representations by predicting the discrete tokens from images and videos. \cite   {MAE} proposed an asymmetric encoder-decoder design for masked image modeling, and \cite   {videomae} adapted this pre-training framework for videos.
Recent works have shown the success of Masked Autoencoders (MAE) in reconstructing time-series data \cite   {ti_mae}, unifying representation learning and image synthesis \cite   {mage_image}, or obtaining useful representations for the robotic domain  \cite   {mae_repr_robots}. In this paper, we propose a novel approach to adapt the image patch decomposition from ViT \cite   {vit} to human poses and formulate the motion synthesis as a reconstruction problem inspired by MAEs.

\section{Methodology}
\begin{figure*}
\centering
\includegraphics[width=0.98\textwidth]{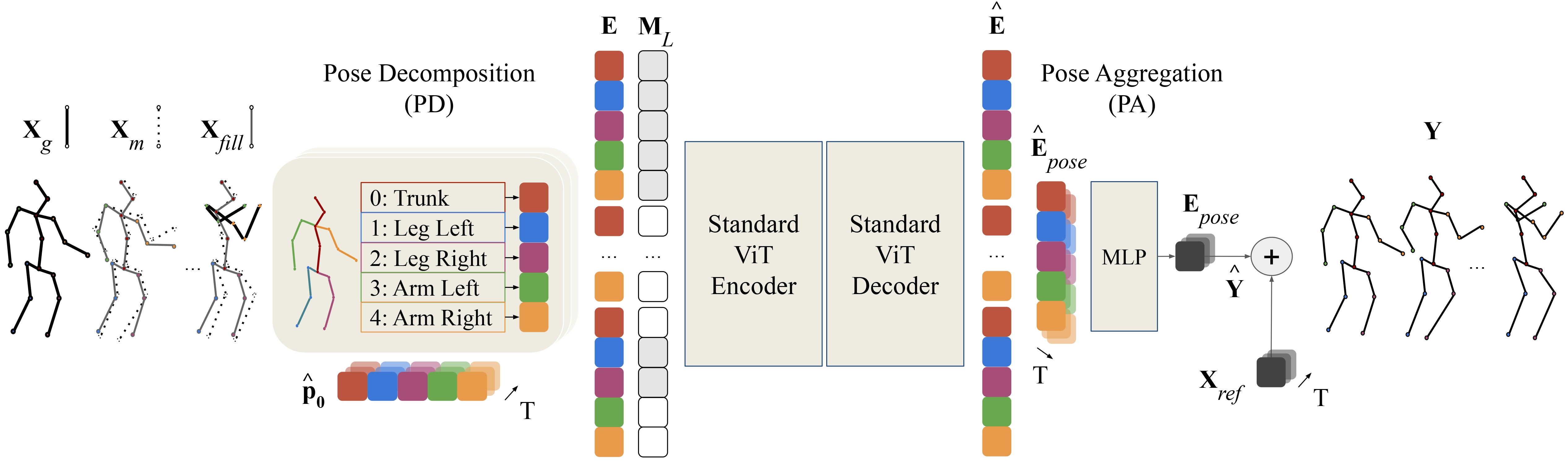}
\caption{\textbf{UNIMASK-M architecture.} Let a human motion  $\mathbf{X}$ and its respective binary mask $M$. We first interpolate $\mathbf{X}_g$ to obtain $\mathbf{X}_{fill}$ and provide consistency to the input. Then, our Pose Decomposition module (PD) deconstructs each pose $\mathbf{p}_t$ into a sequence of patches $\hat{\mathbf{p}}_t$, which we project and flatten to a sequence of tokens $\mathbf{E}$. We add the $emb_{mix}$ to $\mathbf{E}$ to inform the encoder and decoder about the masked tokens and the spatio-temporal structure. Our ViT-based encoder and decoder reconstruct the sequence of tokens. Our Pose Aggregation module (PA) regroups the decoded tokens into poses using an MLP layer. Finally, each pose is projected back to the joint representations and summed to our reference motion $\mathbf{X}_{ref}$.}
\label{fig:architecture}
\end{figure*}
In this section, we present our Unified Masked Autoencoder with patchified SKeletons for Motion synthesis, called UNIMASK-M. First, we reformulate the human motion synthesis task as a reconstruction problem. Then we describe the different components of our approach. A visual illustration of our UNIMASK-M architecture is depicted in Fig. \ref{fig:architecture}.

\textbf{Problem formulation.} Let $\mathbf{p}_t=[p_{t,1},\cdots,p_{t,J}] \in \mathbb{R}^{J\times n}$ be a human pose at time $t$ composed by $J$ joints and $\mathbf{X}=[\mathbf{p}_{0}, \cdots, \mathbf{p}_{T}]  \in \mathbb{R}^{T \times J \times n}$ a human motion. Each joint $p_{t,j}$ can be described in various representations such as the standard euclidean $xyz$-position ($n=3$) or the robust ortho-6D rotation \cite   {ortho6d} ($n=6$). Let also a binary mask $\mathbf{M}=[\mathbf{m}_{0}, \cdots, \mathbf{m}_{T}]\in\mathbb{R}^{T \times J \times n}$ indicate the visibility of each joint in $\mathbf{X}$. The task of motion synthesis is defined as the reconstruction of the missing joints $\mathbf{X}_m= \mathbf{X} \odot \mathbf{M}$ conditioned by the known motion $\mathbf{X}_g = \mathbf{X} \odot (1-\mathbf{M})$.

\textbf{Baseline strategy.}  We adapt the input and output motion to facilitate the model's ability to learn and make accurate predictions, as done in \cite   {2CH-TR, DELTA_INTERP}.  First, we fill the missing joints $\mathbf{X}_m$ using interpolation or repeating the last observed information. We denote this filled motion as $\mathbf{X}_{fill}$, which provides smoothness and continuous flow of motion.  By interpolating the masked motion, we ensure no sudden jumps or disruptions in the motion information, which could negatively affect the model's ability to learn and make accurate predictions. Second, we adopt the delta-strategy from \cite   {DELTA_INTERP}, where the output of our neural network $f_{\theta}(\mathbf{X}_{fill})$ represents the deviation from an interpolated reference motion $\mathbf{X}_{ref}$ to the ground-truth $\mathbf{Y}$. Equation  \ref{eq:x_ref} summarizes this strategy. This approach has the advantage of handling global reference frame shifts while improving the smoothness of the transitions from the key-poses. More information about this strategy is detailed in the Suppl. Materials.

%\vspace*{-3mm}
\begin{equation}
\begin{gathered}
    \mathbf{Y} =  f_{\theta}(\mathbf{X}_{fill}) + \mathbf{X}_{ref}
\end{gathered}
\label{eq:x_ref}
\end{equation}
%\vspace*{-3mm}

\textbf{Pose Decomposition module (PD).} We propose a novel Pose Decomposition module (PD) to deconstruct a sequence of skeletons ($\mathbf{X}_{fill}$) into flattened limb-based patches. We group joints belonging to each limb (legs and arms) and trunk, and restructure the human pose $\mathbf{p}_t$ into $L$ patches $\hat{\mathbf{p}}_t=[\hat{\mathbf{p}}_{t}^0, \cdots, \hat{\mathbf{p}}_{t}^L] $, where each patch $\hat{\mathbf{p}}_t^l$ has a pre-defined number of joints $n_l$. For a given pose $\mathbf{p}_t$, each patch $\hat{\mathbf{p}}_{t}^l$ is projected independently to a token representation  $\mathbf{e}_t^l  \in \mathbb{R}^{D}$ through a linear layer. Then, we flatten all patches for the entire motion, creating a sequence of tokens $\mathbf{E} \in \mathbb{R}^{L \cdot T \times D}$. We also restructure the input mask $\mathbf{M}$ into patches, such that $\mathbf{M}_{L} \in \mathbb{R}^{L \cdot T}$ represents a binary mask per token. For instance, if the binary mask for the token of the trunk has a value of 1, we consider the trunk to be visible. 

The PD module serves two purposes. Firstly, it allows the model to condition the motion only on partial information, the tokens, unlike previous approaches that could only consider the whole pose as a condition. This is because we project patches independently to the representation space instead of projecting the full-body pose. Secondly, by flattening the patches into a sequence of tokens, our UNIMASK-M can leverage self-attention to capture both the spatial and temporal relationships in a single step, contrary to the parallel processing used in \cite{2CH-TR}. We demonstrate the performance boost of using PD compared to the existing full-body projection in our Ablation Study.

\begin{figure}
\centering
\includegraphics[width=0.42\textwidth]{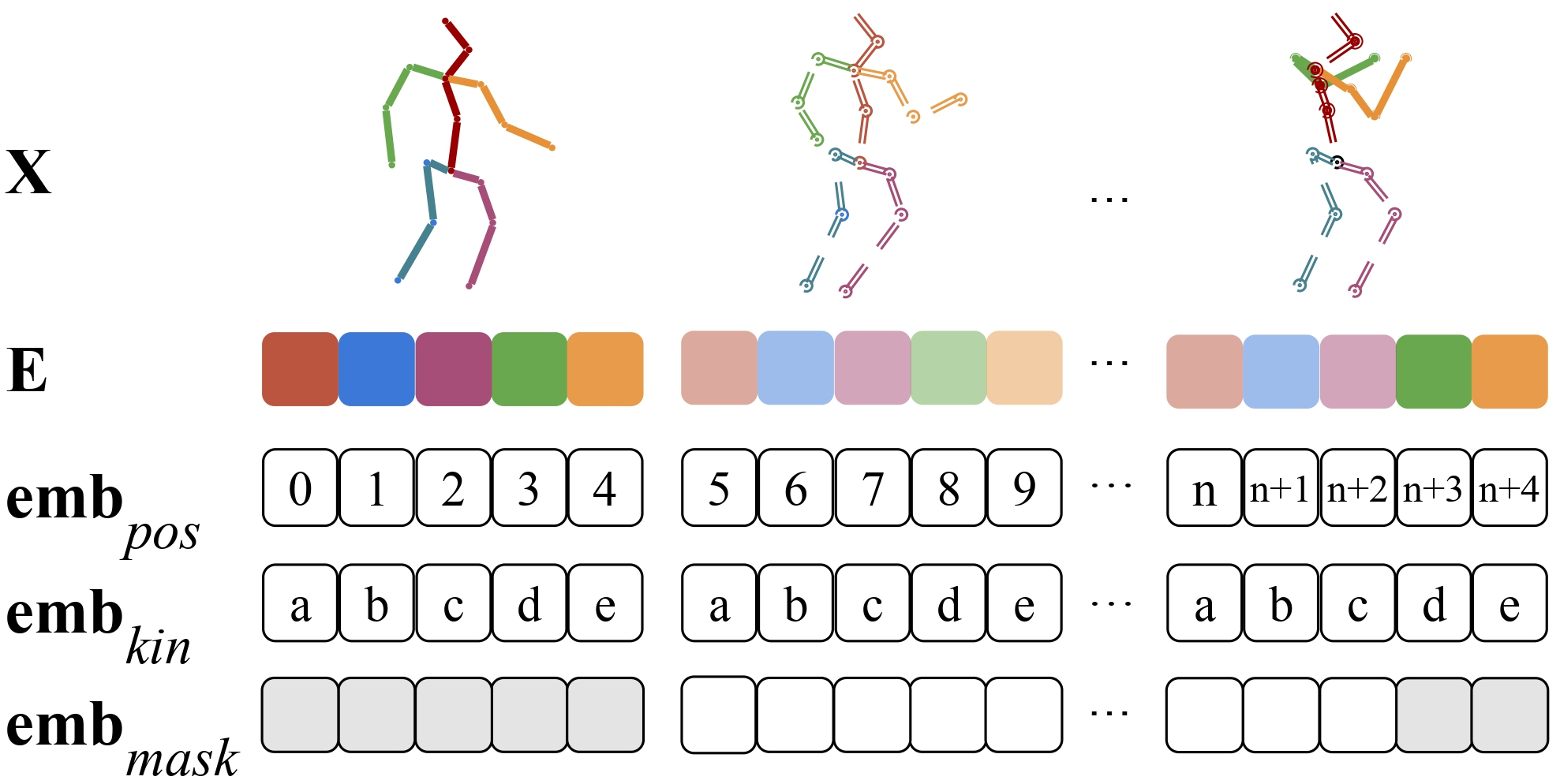}
\caption{\textbf{Mixed embedding strategy.} The mixed embeddings are obtained by summing (i) a masking token to identify the masked patches ($emb_{mask}$); (ii) $L=5$ spatial learnable parameters that correspond to each body part ($emb_{kin}$); and (iii)  a sinusoidal position embedding ($emb_{pos}$).}
\label{fig:mixed_embedding}
\end{figure}

\textbf{Mixed Embeddings.} We adopt a mixed embeddings strategy $emb_{mix}$ to inform the model about the spatio-temporal structure of $\mathbf{X}_{fill}$ and the identification of the masked patches $\mathbf{M}$ to be reconstructed, as shown in Fig. \ref{fig:mixed_embedding}. For that, we extend the sinusoidal positional encoding ($emb_{pos}$) used in Transformers with a kinematic embedding ($emb_{kin}$). $emb_{kin}$ is a patch-dependent learnable parameter added to each token belonging to a particular limb. This embedding informs about the kinematic structure of the input sequence. By incorporating $emb_{kin}$, the model can distinguish between different body parts and leverage their spatial relationships. Finally, inspired by MAEs, we add an  learnable parameter to the missing patches ($emb_{mask}$).

\textbf{Masked Autoencoder}. We adopt the standard ViT \cite   {vit} self-attention encoder and decoder block, composed by Multi-Layer Perceptrons and Layer Normalization \cite   {layernorm}. Unlike previous works \cite   {MAE, videomae, mage_image}, we do not neglect the masked patches in the encoder phase, as all patches $\mathbf{E}$ have relevant information for the model (due to the adaptation of $\mathbf{X}$ into $\mathbf{X}_{fill}$). We verify our hypothesis thought an Ablation Study. Finally, both the encoder and the decoder share the same $emb_{mix}$.

\textbf{Pose Aggregation module (PA)}. The output of the decoder is a flattened sequence of reconstructed tokens denoted by $\mathbf{\hat{E}} \in \mathbb{R}^{L\cdot T \times D}$. To leverage the hierarchical structure of a human skeleton, we re-group the tokens that correspond to a particular pose into $\mathbf{\hat{E}_{pose}} \in \mathbb{R}^{T \times L\cdot D}$ and then project it to a single pose representation $\mathbf{E_{pose}} \in \mathbb{R}^{T \times D}$ using two fully connected layers with GELU activation \cite   {gelu}. PA allows fusing the different body-part token representations into a full-body pose, which facilitates the prediction of more plausible motions. Finally, we project $\mathbf{E_{pose}}$ to the joint-based pose $\hat{\mathbf{Y}} \in \mathbb{R}^{T \times P}$ and add it to the reference motion $\mathbf{X}_{ref}$.

\section{Experimental results}
In this section, we evaluate the effectiveness of our UNIMASK-M model across different tasks and datasets by conducting multiple experiments and comparisons. Unless stated otherwise, all models in comparison were trained for a target task within the proposed dataset. More visual results and implementation details in the Suppl. Materials.

\textbf{Dataset and Metrics}. Human motion forecasting has been mainly addressed in Human3.6M dataset \cite   {h36m_pami}. This dataset includes $3.6$ million 3D poses of humans performing 15 daily activities. Following standard evaluation, we report Mean Per Joint Position Error (MPJPE) in the test subject S5, as suggested in \cite   {RNN_Motion2, DCT-GCN, mao2020history} when predicting the next 1s motion given 400ms of the past. Human motion inbetweening is evaluated in LaFAN1 dataset \cite   {lafan1dataset}. This dataset contains $496,672$ motions sampled at 30Hz and captured in a MOCAP studio.  We report the average L2 distance of the global position (L2P) or rotation (L2Q, in quaternion) per frame, as well as the Normalized Power Spectrum Similarity (NPSS) to evaluate angular differences between prediction and ground truth on the frequency domain. The given key-poses consists of of 10 poses from the past and 1 pose from the future as input. The number of intermediate poses to be predicted can vary between 5 and 30. These values were proposed by previous works \cite   {TRANSFORMER_MIB1, DELTA_INTERP, RNN2_MIB}.

\subsection{Quantitative evaluation}

\begin{figure}
\centering
\includegraphics[width=0.45\textwidth]{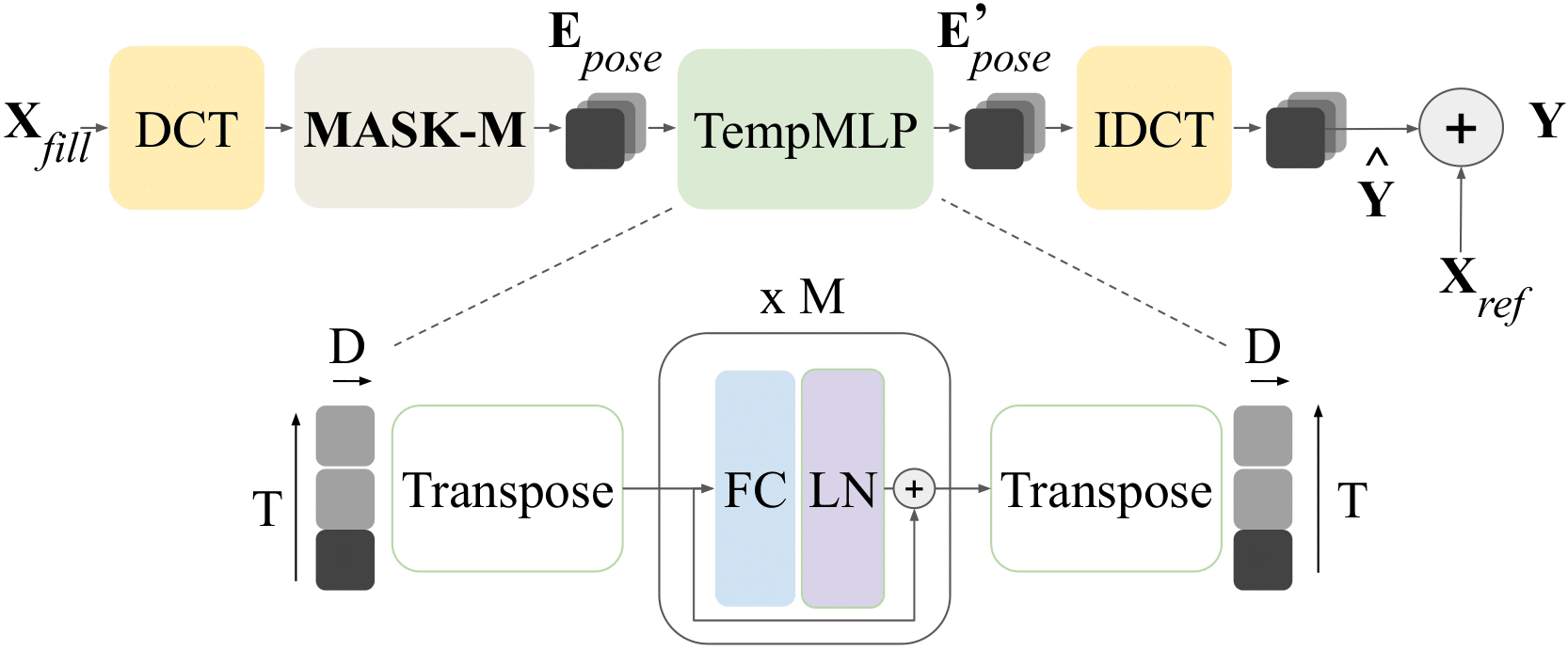}
\caption{\textbf{Adaption of UNIMASK-M using DCT and TempMLP.} First, we apply Discrete Cosine Transformation (DCT) and Inverted DCT (IDCT) to encode and decode the given motion. Additionally, we adopt a Temporal MLP (TempMLP) module to refine the predicted pose sequence through $M$ blocks of fully connected layers (FC), Layer Normalization (LN) and a residual connection. }
\label{fig:dct_tempmlp}
\end{figure}

\begin{figure}
\centering
\includegraphics[width=0.47\textwidth]{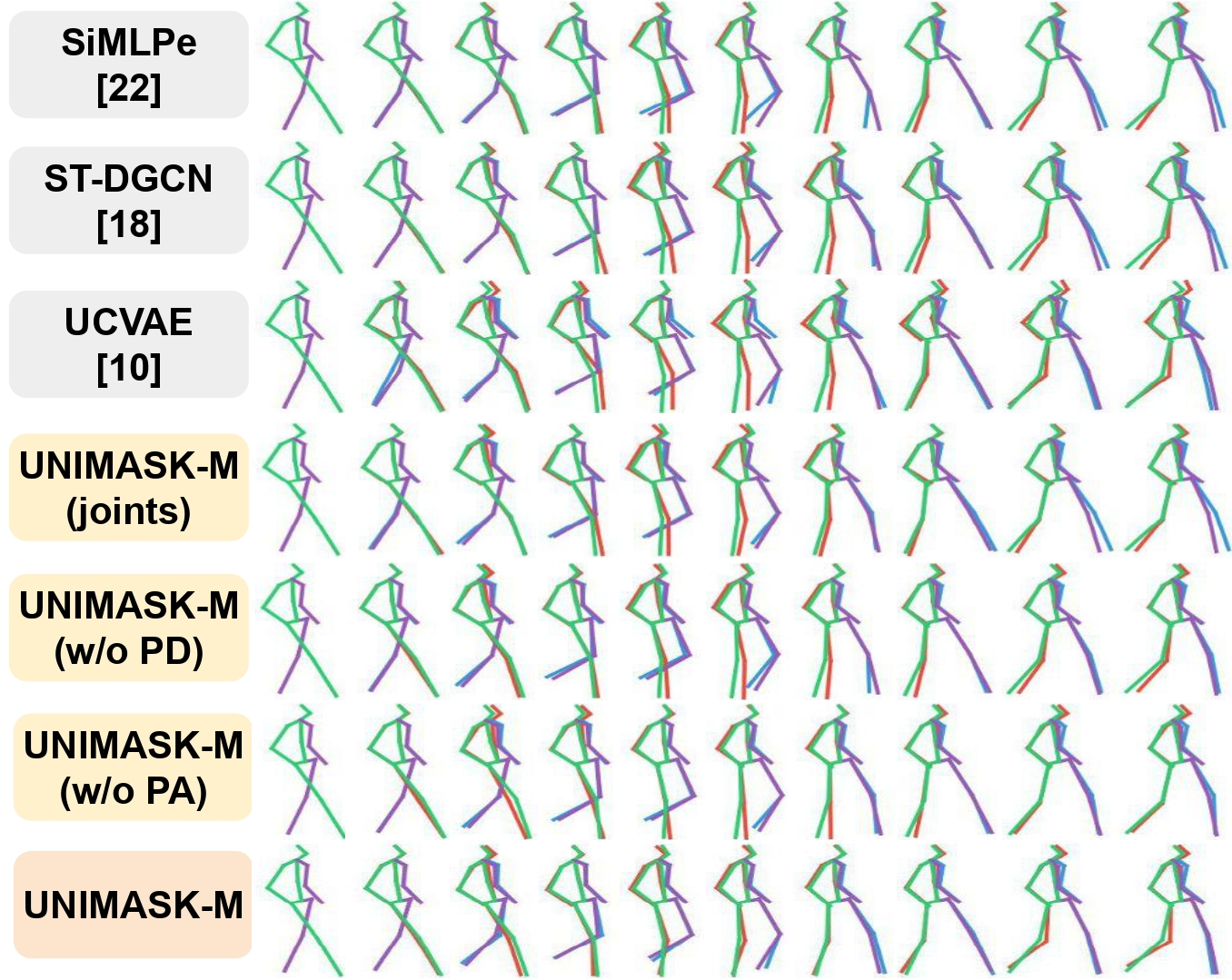}
\caption{\textbf{Comparison of the motion forecasting task.} Predicted skeletons are shown in red and blue.}
\label{fig:baselinevisualcomparison}
%\vspace{-1mm}
\end{figure}

\textbf{Human Motion Forecasting.} Our model is tested in Human3.6M dataset and shows competitive results to the state-of-the art models, as shown in Table \ref{tab:quantitative_forecasting}. In addition to the UNIMASK-M architecture, we adopt the Discrete Cosine Transformation (DCT) to the input motion due to the periodic movements accounted for in this dataset. This strategy is also adopted for most works shown in Table \ref{tab:quantitative_forecasting} as it boosts the performance  \cite   {DCT-GCN, ST-DGCN, MSR-GCN, SIMPLE}. Moreover, we smooth the transition from prefix sequence to predicted poses using an additional Temporal MLP (TempMLP) that refines the decoded poses $\mathbf{E}_{pose}$ in time. These modifications, shown in Fig. \ref{fig:dct_tempmlp}, are adopted only for Human3.6M due to the cyclic motions encountered. We also provide our results without them in Table \ref{tab:quantitative_forecasting}. It should be noted that to predict future poses using SiMLPe \cite   {SIMPLE} and \cite   {mao2020history}, it is necessary to observe a sequence of the past that is five times longer. Moreover, both SiMLPe and ST-DGCN \cite   {ST-DGCN} are autoregressive models. On the contrary, our UNIMASK-M model observes shorter sequences and predicts the whole motion in one forward pass. A visual comparison is depicted in Figure \ref{fig:baselinevisualcomparison}.

\begin{table}
\centering
\resizebox{0.47\textwidth}{!}{%
\begin{tabular}{@{}lcccccc@{}}
milliseconds (ms) & \textbf{80}   & \textbf{160}        & \textbf{320}        & \textbf{400}        & \textbf{560}        & \textbf{1000}  \\ \midrule
Repeat last pose  & 23.8 & 44.4       & 76.1       & 88.2       & 107.4      & 136.6 \\ \midrule
\cite {MotionMixer}          & 12.8 & 26.6       & 53.1       & 64.5       & 82.9       & 117.1 \\
\cite {MSR-GCN}            & 12.1 & 25.5       & 51.6       & 62.9       & -          & 114.2 \\
\cite {mao2020history}      & 10.4 & 22.6       & 47.1       & 58.3       & 77.3       & 112.1 \\
\cite {DCT-GCN}           & 12.4 & 25.2       & 49.9       & 60.9       & 79.5       & 112.7 \\
SiMLPe \cite {SIMPLE}            & \textbf{9.7}  & \textbf{22.0}       & 47.0       & 58.1       & 76.7       & 110.5 \\
ST-DGCN \cite {ST-DGCN}           & 10.3 & 22.6       & \textbf{46.6}       & \textbf{57.5}       & \textbf{76.3}       & \textbf{110.0} \\ \midrule
UNIMASK-M (baseline)      & 17.7 & 33.3 &  59.8 & 70,8 & 89,0 & 120,6  \\
UNIMASK-M (w/ DCT)      & 15.8 & 29.1 &  53.8 & 64.4 & 81.7 & 113.6  \\
UNIMASK-M (Ours)      & {  11.9} &  {  25.1} &   {  50.7} &  {  61.6} &  {  79.6} &  {  112.1}  \\  \bottomrule
\end{tabular}%
}
\caption{Quantitative comparison of MPJPE error in 3D human motion forecasting for Human3.6M dataset. Here, bold denotes the best result at each time-horizon.}
\label{tab:quantitative_forecasting}
\end{table}

On top of that, we also evaluate the motion forecasting performance when the input motion contains occlusions. For that, we randomly mask the joints for the observed sequence with a probability $p$. For instance, $p=0.2$ implies that $20\%$ of the joints are not visible. We apply the same approach to pure forecasting models from literature and show their results in Table \ref{tab:quantitative_fc_occlusion}. Results show that our UNIMASK-M is able to better handle occlusions in the input sequence.  

\begin{table}
\centering
\resizebox{0.47\textwidth}{!}{%
\begin{tabular}{@{}lcccccc@{}}
milliseconds (ms) & \textbf{80}   & \textbf{160}        & \textbf{320}        & \textbf{400}        & \textbf{560}        & \textbf{1000}  \\ \midrule
\cite {MotionMixer}        & 32.7 & 50.1 & 76.9 & 86.8 & 102.0 & 129.1 \\
\cite {mao2020history}      & 32.4 & 48.8 & 72.9 & 82.0 & 96.3  & 125.3          \\
\cite {DCT-GCN}            & 32.8 & 49.8 & 74.4 & 83.5 & 97.8  & 125.8          \\
SiMLPe \cite {SIMPLE}           & 32.7 & 49.6 & 74.0 & 82.8 & 96.7  & 124.7          \\
ST-DGCN \cite {ST-DGCN}           & 32.8 & 49.5 & 74.3 & 83.5 & 97.8  & 124.6          \\ \midrule
UNIMASK-M (Ours)      & \textbf{27.3} & \textbf{41.5} & \textbf{64.9} & \textbf{74.5} & \textbf{90.3} & \textbf{120.5}  \\  \bottomrule
\end{tabular}%
}
\caption{Quantitative comparison of MPJPE error in 3D human motion forecasting when the observed sequence has 20\% joints occluded in Human3.6M dataset.}
\label{tab:quantitative_fc_occlusion}
\end{table}

\begin{figure}
\centering
\includegraphics[width=0.47\textwidth]{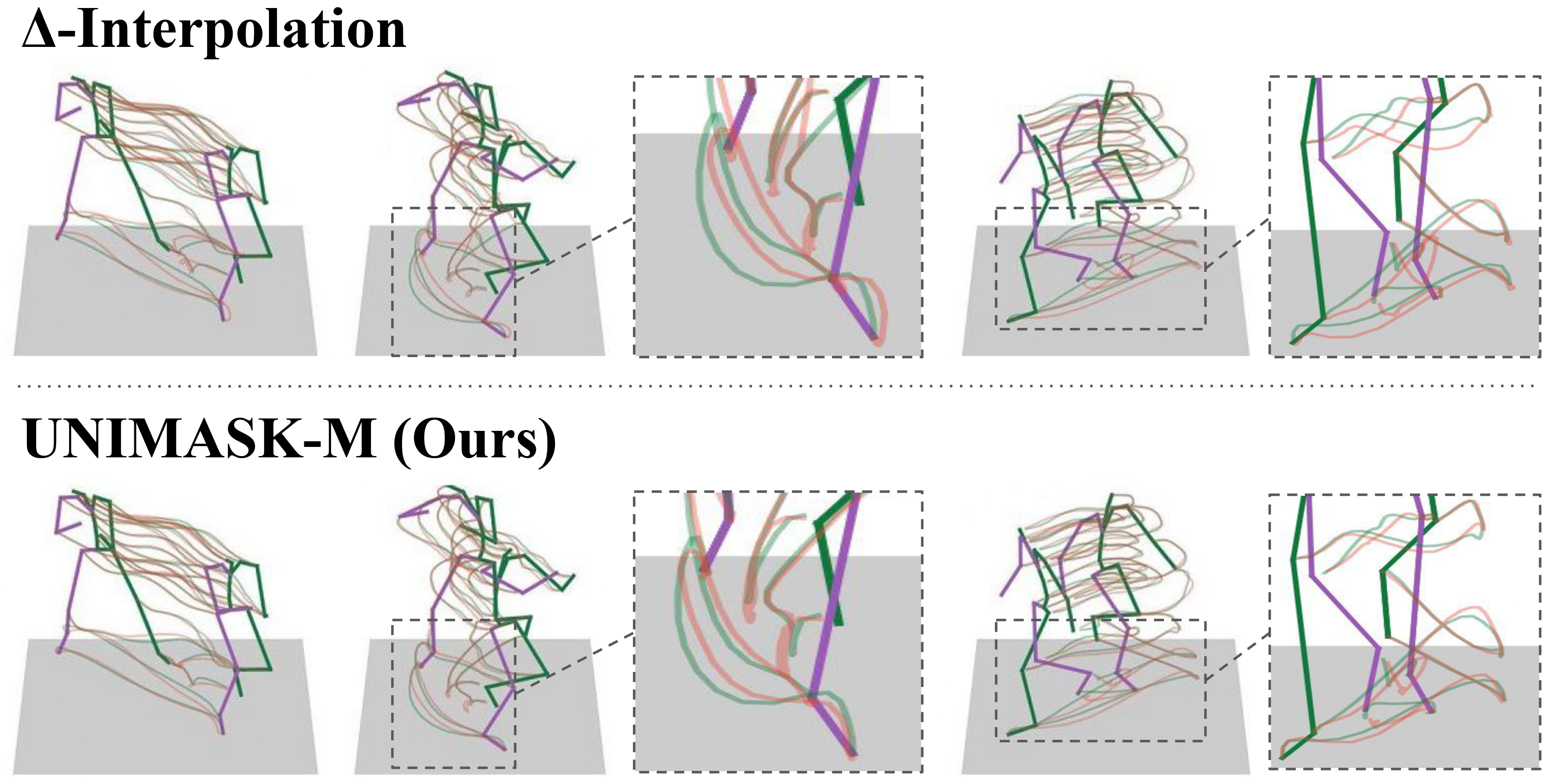}
\caption{\textbf{Visual comparison of the inbetweening task with \cite{DELTA_INTERP} }. We show the predicted motion trace for both \cite   {DELTA_INTERP} (top row) and our UNIMASK-M (bottom row). The results show that our predicted trace (red) is closer to the ground-truth trace (green).}
\label{fig:lafan1_qualitative}
\end{figure}

\begin{figure}
\centering
\includegraphics[width=0.47\textwidth]{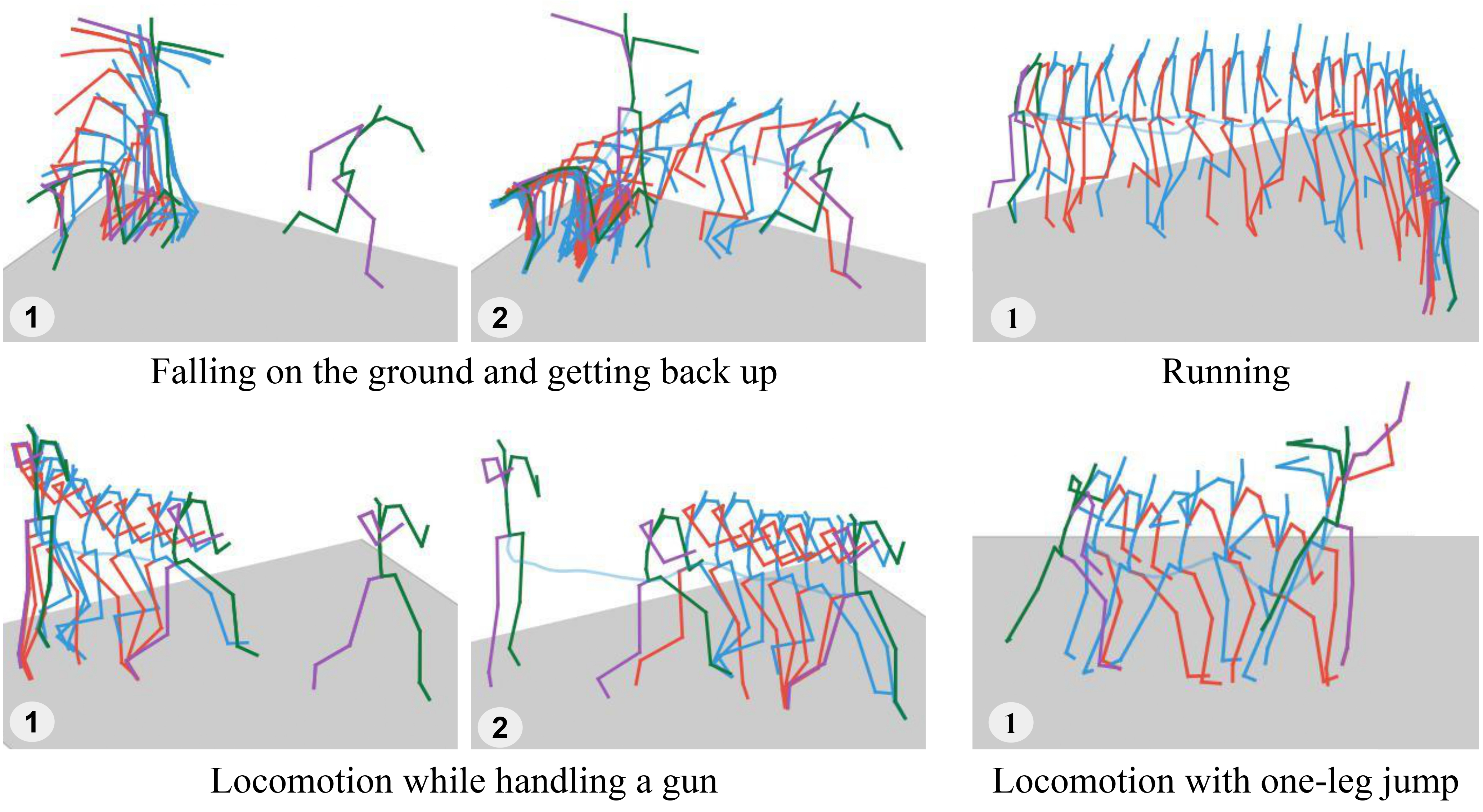}
\caption{\textbf{Qualitative evaluation of human motion inbetweening in LaFAN1 Dataset}. Given certain key-poses (green and purple skeletons) our UNIMASK-M fills the motion in-between (red and blue skeletons).}
\label{fig:lafan1_qualitative_vis}
\end{figure}

\textbf{Human Motion Inbetweening.} We evaluate our UNIMASK-M model for motion inbetweening in LaFAN1 dataset, and show the results in Table \ref{tab:lafan1_quantitative_all}. Our UNIMASK-M performs better for longer transitions and with similar results in shorter periods than the state-of-the-art \cite   {DELTA_INTERP}. \cite   {DELTA_INTERP} decouples the prediction using cross-attention for the masked frames and self-attention for the observed poses, facilitating the prediction of the transition close to the key-poses. However, our encoder-decoder structure provides more robustness in longer transitions and therefore is more appropriate for synthesis tasks. To confirm this claim, we additionally train both  \cite   {DELTA_INTERP} and UNIMASK-M when predicting longer transitions (50 and 70). These results are reported in Table \ref{tab:lafan1_quantitative_all} and show how our UNIMASK-M clearly outperforms state-of-the-art for long-term motion inbetweening. Note that these trained long-term models also perform in-betweening for any transition frames. Qualitative results are observed in Figure \ref{fig:lafan1_qualitative} and \ref{fig:lafan1_qualitative_vis}.

\begin{table*}[]
\centering
\resizebox{\textwidth}{!}{%
\begin{tabular}{@{}lccccccccccccccc@{}}
  & \multicolumn{5}{c}{\textbf{L2Q $\downarrow$}}              & \multicolumn{5}{c}{\textbf{L2P $\downarrow$}}              & \multicolumn{5}{c}{\textbf{NPSS $\downarrow$}}                   \\ \cmidrule(lr){2-6}  \cmidrule(lr){7-11} \cmidrule(lr){12-16}
Number of Transition Frames & \textbf{5}    & \textbf{15}   & \textbf{30}   & \textbf{50}  & \textbf{70}  & \textbf{5}    & \textbf{15}   & \textbf{30}  & \textbf{50}  & \textbf{70}   & \textbf{5}      & \textbf{15}     & \textbf{30}    & \textbf{50}  & \textbf{70}  \\ \midrule
Zero velocity         & 0.56 & 1.10 & 1.51 & 1.90 & 2.22 & 1.52 & 3.69 & 6.60 & 6.31 & 8.16 & 0.0053 & 0.0522 & 0.2318 & 0.6009 & 1.2194 \\
SLERP Interpolation   & 0.22 & 0.62 & 0.98 & 1.34 & 1.72 & 0.37 & 1.25 & 2.32 & 2.51 & 3.54 & 0.0023 & 0.0391 & 0.2013 & 0.5891 & 1.2385 \\ \midrule
TG$_{rec}$ \cite {lafan1dataset}               & 0.21 & 0.48 & 0.83 & - & - & 0.32 & 0.85 & 1.82 & - & - & 0.0025 & 0.0304 & 0.1608 & - & - \\
TG$_{complete}$ \cite {lafan1dataset}           & 0.17 & 0.42 & 0.69 & - & - & 0.23 & 0.65 & 1.28 & - & - & 0.0020 & 0.0258 & 0.1328 & - & - \\
SSMCT$_{local}$   \cite {TRANSFORMER_MIB1}         & 0.17 & 0.44 & 0.71& - & -  & 0.23 & 0.74 & 1.37 & - & - & 0.0019 & 0.0291 & 0.1430 & - & - \\
SSMCT$_{global}$  \cite {TRANSFORMER_MIB1}         & 0.14 & 0.36 & 0.61 & - & - & 0.22 & 0.56 & 1.10 & - & - & 0.0016 & 0.0234 & 0.1222 & - & - \\
$\Delta$-Interpolation \cite {DELTA_INTERP}  & \textbf{0.11} & \textbf{0.32} & \textbf{0.57} & {  0.87} & {  1.20} & \textbf{0.13} & {  0.47}    & {  1.00}   & {  1.28} & {  1.91} & \textbf{0.0014} & {  0.0217}    & {  0.1215}   & {  0.4042} & {  1.0180} \\
%Tween Transformer     & 0.16 & 0.39 & 0.65 & 0.21 & 0.59 & 1.21 & 0.0019 & 0.0261 & 0.1358 \\
UNIMASK-M (Ours)    & {  0.12}    & \textbf{0.32} & \textbf{0.57} & \textbf{0.83} & \textbf{1.16} & {  0.14}    & \textbf{0.46} & \textbf{0.97} & \textbf{1.20} & \textbf{1.80} &\textbf{0.0014} & \textbf{0.0216} & \textbf{0.1204} & \textbf{0.3716} & \textbf{0.9532}\\ \bottomrule
\end{tabular}%
}
\caption{\textbf{Quantitative evaluation of human motion inbetweening on LAFAN1 dataset}. A lower score is better. Here, bold indicates the best result. Note that we trained a different model for the 50 and 70 transition frames for both \cite {DELTA_INTERP} and ours}
\label{tab:lafan1_quantitative_all}
\end{table*}

\textbf{Human Motion Completion.} We assess the motion completion task in the Human3.6M dataset by introducing random occlusions throughout the entire motion at varying percentages. Since there is no established benchmark for this topic, we adapt and train \cite   {DELTA_INTERP} model specifically for this purpose, which we call CrossViT. This baseline CrossViT uses our PD module to allow the conditioning based only on body-parts.  We evaluate the motion completion task when masking 90\% of future joints in the Human3.6M dataset and observe that our UNIMASK-M reduces the MPJPE error by -$12.54\%$ compared to CrossViT. %For a quantitative evaluation of our approach, we refer to Table \ref{tab:quantitative_completion}, which includes results obtained by masking 90\% of future joints in the Human3.6M dataset. These results show the advantage of our proposed UNIMASK-M in the motion completion task.

\textbf{Unified model for different tasks}. Although our previous results demonstrate the versatility of our model for various tasks, we train a unified model with different masking patterns to achieve high performance in several tasks at once. Additionally, we implement curriculum learning,  which gradually increases the probability of masking during training to enhance the model's robustness in challenging tasks. The advantage of using curriculum learning in our training is shown in Table \ref{tab:curricular learning}, where the model exhibits performance across tasks comparable to task-specific models.

\textbf{Comparison with generative models}. We trained the generative U-CVAE \cite{UnifiedCVAE} with their official code and measure the Top1 MPJPE error from $30$ synthesized motions. Our UNIMASK-M significantly outperforms U-CVAE in both motion forecasting (+$64.49\%$) and completion (+$88.12\%$). We compared qualitatively both UNIMASK-M and U-CVAE in Fig. \ref{fig:baselinevisualcomparison}. Motions from U-CVAE are jittery and tend to collapse in the long term. Moreover, U-CVAE is trained only with whole pose masking and generates non-natural behaviors when given partial poses. Finally, MDM \cite{MDM} does not handle key poses and lacks real-time capabilities. We compute the average run time per 10000 runs and observe that our proposed UNIMASK-M is  $494.79\times$ faster than MDM. %as shown in Table \ref{tab:computationalresources}. 

\begin{table}[]
\centering
\resizebox{0.43\textwidth}{!}{%
\begin{tabular}{lcccccc}
  & \multicolumn{6}{c}{\textbf{Motion Forecasting (MPJPE) }$\downarrow$}         \\
  \midrule
\textbf{Mask Probability} & \multicolumn{1}{l}{\textbf{80}}   & \textbf{160}  & \textbf{320}  & \textbf{400}  & \textbf{560}  & \textbf{1000}          \\
\midrule
$1$                       & \multicolumn{1}{l}{\textbf{11.9}} & 25.1          & 50.7          & 61.6          & \textbf{79.6} & \textbf{112.1}         \\
$0.9$                     & \multicolumn{1}{l}{-}             & -             & -             & -             & -             & -             \\
$0.85 \rightarrow 1$      & \multicolumn{1}{l}{\textbf{11.9}} & \textbf{24.6} & \textbf{49.9} & \textbf{61.2} & 80.2          & 115.7         \\

\midrule
\multicolumn{1}{c}{}     & \multicolumn{6}{c}{ \textbf{Motion Completion (MPJPE)} $\downarrow$}  \\
  \midrule
\textbf{Mask Probability} & \textbf{80}                       & \textbf{160}  & \textbf{320}  & \textbf{400}  & \textbf{560}  & \textbf{1000} \\
\midrule

$1$                       & 13.0                              & 23.7          & 41.2          & 47.5          & 56.9          & 82.2          \\
$0.9$                     & \textbf{8.4}                      & \textbf{14.2} & \textbf{22.4} & \textbf{25.1} & \textbf{29.3} & 52.0          \\
$0.85 \rightarrow 1$      & 8.5                               & 14.7          & 22.9          & 25.7          & 29.9          & \textbf{51.9}
\end{tabular}%
}
\caption{\textbf{MPJPE error of UNIMASK-M in Human3.6M dataset under different training masking probabilities ($p_m$) and tasks.} Training on forecasting requires the whole future mask ($p_m=1$) while training on completion uses random patch masking ($p_m=0.9$). To boost performance in both objectives, curriculum learning is applied where $p_m$ increases during training epochs (from $p_m=0.85$ to $p_m=1$).}
\label{tab:curricular learning}
\end{table}
%\input{tables/computational_resources}

%\subsection{Qualitative evaluation}

\subsection{Ablation Study}
\label{sec:ablation_study}
This section analyses the various approaches proposed in this work and their impact on performance. Table \ref{tab:lafan1_ablation} shows the results of the ablation study.

\textbf{Pose decomposition (PD)}. Our UNIMASK-M proposes PD to leverage self-attention among body parts but aggregates the spatial information prior to the pose projection through PD. We compare our PD approach against the typical linear projection proposed in the literature for pose encoding. We also design an inverted PD that linearly projects each decoded body parts patch  $\mathbf{\hat{E}}$ to the assigned joints independently, and we compare it with our PA approach.  Table \ref{tab:lafan1_ablation} shows the performance boost of our design.

\textbf{DCT and TempMLP}. We evaluate our model with DCT and Temporal MLPs, as depicted in Fig. \ref{fig:dct_tempmlp}, and observe that DCT highly degrades the model's performance. Note that the motions from the LaFAN1 dataset are much more complex and less periodical, therefore resulting in a loss of information when working on the frequency domain obtained from DCT. On the other hand, plugging an additional Temporal MLP causes the model to overfit.

\begin{table}[]
\centering
\resizebox{0.47\textwidth}{!}{%
\begin{tabular}{@{}lcccccc@{}}
  & \multicolumn{2}{c}{\textbf{L2Q $\downarrow$}} & \multicolumn{2}{c}{\textbf{L2P $\downarrow$}} & \multicolumn{2}{c}{\textbf{NPSS $\downarrow$}} \\ \cmidrule(lr){2-3}  \cmidrule(lr){4-5} \cmidrule(lr){6-7}
Num Frames & \textbf{5}    & \textbf{30}   & \textbf{5}  & \textbf{30}   & \textbf{5}      & \textbf{30}     \\ \midrule
w/o PD                                                           & 0.13                   & 0.59          & 0.15                 & 1.03           & 0.0015          & 0.1263          \\ 
w/o PA & 0.13                      & 0.60                & 0.15                      & 1.05                & 0.0015                       & 0.1271               \\
\midrule
w/ DCT                                                             & 0.17                                   & 0.68          & 0.23               & 1.33           & 0.0019              & 0.1503          \\
w/ TempMLP & \textbf{0.12} &  0.60 & \textbf{0.14}  & 1.04 & \textbf{0.0014}              & 0.1280           \\ \midrule
w/o $emb_{kin}$                                                & 0.13                           & 0.58          & 0.15                 & 0.98           & \textbf{0.0014}                & 0.1216          \\
w/ Attn Mask                                                    & 0.13                                & 0.59          & 0.15           & 1.02           & 0.0015                & 0.1243          \\ \midrule
w/o encoder & \textbf{0.12}  & 0.58 &\textbf{ 0.14} & 0.99 & \textbf{0.0014} & 0.1248 \\
w/o decoder                                                        & 0.13                                   & 0.60          & 0.16                & 1.05           & 0.0015           & 0.1282          \\
UNIMASK-M Light                                                        & 0.12                                 & 0.59          & \textbf{0.14}                 & 1.00           & \textbf{0.0014}             & 0.1248          \\ \midrule
\textbf{UNIMASK-M} & \multicolumn{1}{l}{{\textbf{0.12}}}  & \textbf{0.57} & \textbf{0.14}  & \textbf{0.97}  & \textbf{0.0014} & \textbf{0.1204} \\ \hline
\end{tabular}%
}
\caption{Performance of our UNIMASK-M under different configurations in the inbetweening task and the LaFan1 dataset \cite{lafan1dataset}. }
\label{tab:lafan1_ablation}
\end{table}

\textbf{Attention mechanism}. To rely on the self-attention for both masked and key-poses, our model takes advantage of the mixed embeddings to be informed about (i) the temporal relations, (ii) the masked poses, and (iii) the spatial relations. While (i) and (ii) are already proposed for motion synthesis \cite   {TRANSFORMER_MIB1}, (iii) is considered a novelty due to the uniqueness of the PD. We show the importance of using $emb_{kin}$ in Table \ref{tab:lafan1_ablation}.
We also train our model with a triangular attention mask to enforce causality. As expected due to the bi-directionality of the inbetweening tasks, imposing causal attention degrades performance.

\textbf{Model size}. The final UNIMASK-M proposed for motion inbetweening in LaFAN1 dataset has 41.4M parameters, with 3M parameters less than \cite   {DELTA_INTERP}. Inspired by MAEs, we evaluate our model using different configurations of depth and width in the encoder-decoder structure. We observe that changing the depth of the model does not have a high impact on the performance while decreasing the width has a direct effect on the motion quality for longer transitions. Still, our \textit{UNIMASK-M light} performs better than most previous approaches while having only 12.4M parameters. We also remove the encoder or decoder to reduce the model size, which causes performance degradation. Our UNIMASK-M still exhibits high performance only when using the decoder.

\textbf{Effect of masking ratio.} We analyze different masking ratios for our UNIMASK-M in the motion forecasting with occlusions (i.e. Table. \ref{tab:quantitative_fc_occlusion}). A masking ratio over 50\% increases the MPJPE short-term error by $2.65$, but has a marginal impact in the long term ($\times 1.05$). These findings suggest a strong correlation between the short-term poses and the observed motion, highlighting the vulnerability of auto-regressive approaches \cite{SIMPLE, ST-DGCN} to occlusions, where their MPJPE error increases by $1.19$ and $1.14$ with 40\% occlusion respect to UNIMASK-M.

%%%%%%%%%%%%%%%%%%%%%%%%%%%%%%%%%%%%%%%%%%%%%%%%%%%%%%%%%%%%%%%%%%%%%%%%%%%%%
% TO BE REMOVED IF NOT CONSIDERED APPROPRIATE
% Remember to remove color package as it is not allowed in the code

\textbf{Curriculum learning vs Pretraining.} Some prior Masked-Autoencoders (MAEs) works benefit from an initial pre-training stage, and later finetuning for downstream tasks. We tested this approach and pretrained our UNIMASK-M using random masks for different motion synthesis subtasks and later finetune for motion completion. The results shows a higher MPJPE error (+$0.89$ on average) than the curriculum learning shown in Table \ref{tab:curricular learning}.

\begin{figure}
\centering
\includegraphics[width=0.48\textwidth]{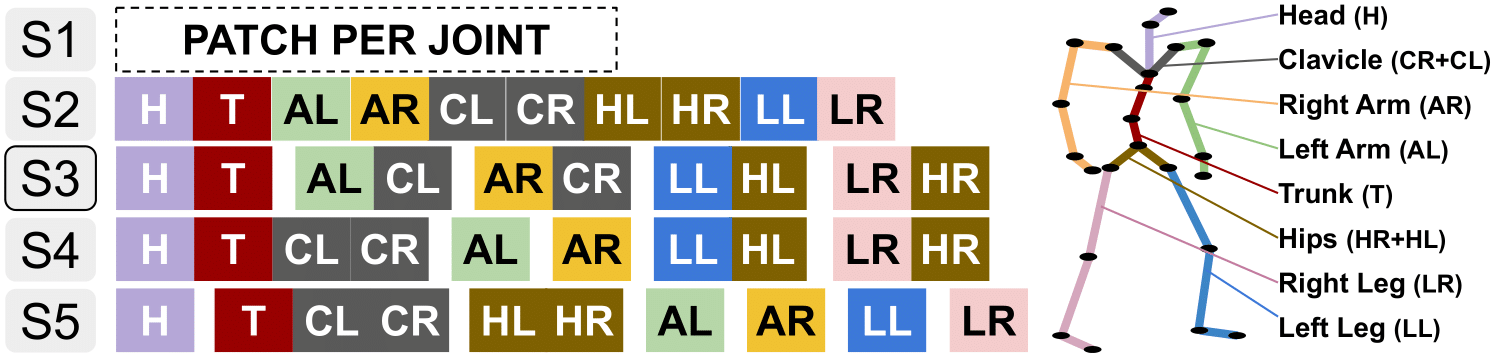}
\vspace{-0.7cm}
%\caption{\textbf{Different patch granularity strategies in Human3.6M skeleton.} }
\label{fig:patch_granularity}
\end{figure}

\begin{figure}[]
\centering
\resizebox{0.48\textwidth}{!}{%
\begin{tabular}{lcccccc}
 \toprule
\textbf{Strategy} & \textbf{80} & \textbf{160} & \textbf{320} & \textbf{400} & \textbf{560} & \textbf{1000} \\  \midrule
S1 (joints) & 12.44 & 25.93 & 51.52 & 62.59 & 81.02 & 114.06 \\
S2 (body, w/o PD) & 12.38 & 25.92 & 51.57 & 62.62 & 80.60 & 113.05 \\
\textbf{S3 (Proposed)} & 11.85 & 25.11 &  {50.69} &  {61.57} & \textbf{79.58 }& \textbf{112.10} \\
S4 & 11.79 & 25.30 & 51.08 & 62.28 & 80.87 & 113.62 \\
S5 & \textbf{11.53} & \textbf{24.78 }& \textbf{50.24} &\textbf{ 61.38} &  {79.86} & 113.38 \\
%S6 &  {11.71} &  {24.93} & 50.58 & 61.78 & 80.54 & 113.31 \\ 
\end{tabular}
}
\vspace{0.05cm}
\caption{MPJPE millimeter error of the motion forecasting task of UNIMASK-M under different patch granularity.}
\vspace{-4mm}
\label{tab:patch_granularity}
\end{figure}

\textbf{Granularity of the patch hierarchy}.  We trained different patch granularities and show the most relevant in Fig. \ref{tab:patch_granularity} for motion forecasting. The results indicate a superior performance on average of our proposed decomposition [S3], with better performance in long-term forecasting which is crucial for real-world applications. Extremely high [S1] and low [S2] granularities underperform, reinforcing the motivation behind our work. Additionally, considering the clavicle as part of the trunk [S4] or the hips as a separate patch [S5] improves short-term predictions, although the motion collapses in the long term. Visually, having a larger trunk  [S2, S4, S5] produces slower movements. In contrast, higher granularity [S1] produces longer sequences of patches, thus increasing the inference time ($\times 1.19\%$) and hindering the model's ability to capture the spatiotemporal relationships, often resulting in a collapse to a certain pose. These behaviors are depicted in Fig. \ref{fig:baselinevisualcomparison} for UNIMASK-M (w/o PD) and UNIMASK-M (joints).

\section{Conclusion}
In this paper, we proposed a unified Masked Autoencoder with patchified SKeletons for Motion synthesis (UNIMASK-M). By considering all motion synthesis subtasks as a masked reconstruction problem, we design a task-agnostic model that can achieve competitive performance compared to the task-dependent state-of-the-art. Ours is the first work to adopt masked autoencoders for general motion synthesis, which might serve as a basis for future works in the field. In addition, we confirm our initial hypothesis that decomposing a human pose into patches provides not only more flexible pose conditioning but also boosts performance. UNIMASK-M shows state-of-the-art results in motion inbetweening for LaFAN1 dataset. Additionally, we obtain competitive performance in motion forecasting in Human3.6M dataset while predicting the whole motion in one-shot.  We also demonstrate the efficiency of UNIMASK-M design for synthesizing robust motions in occluded conditions or when only conditioning on certain body-parts. In conclusion, our approach demonstrates its effectiveness and versatility in generating high-quality human motion predictions in  tasks. 

%\newpage
\bibliography{aaai24}

\begin{thebibliography}{40}
\providecommand{\natexlab}[1]{#1}

\bibitem[{Ahn, Valls~Mascaro, and Lee(2023)}]{DiffPred:2023}
Ahn, H.; Valls~Mascaro, E.; and Lee, D. 2023.
\newblock Can We Use Diffusion Probabilistic Models for 3D Motion Prediction?
\newblock In \emph{2023 IEEE International Conference on Robotics and
  Automation (ICRA)}.

\bibitem[{Aksan et~al.(2021)Aksan, Kaufmann, Cao, and Hilliges}]{ST-TR}
Aksan, E.; Kaufmann, M.; Cao, P.; and Hilliges, O. 2021.
\newblock A spatio-temporal transformer for 3d human motion prediction.
\newblock In \emph{International Conference on 3D Vision (3DV)}, 565--574.
  IEEE.

\bibitem[{Ba, Kiros, and Hinton(2016)}]{layernorm}
Ba, J.~L.; Kiros, J.~R.; and Hinton, G.~E. 2016.
\newblock Layer normalization.
\newblock \emph{arXiv preprint arXiv:1607.06450}.

\bibitem[{Bao, Dong, and Wei(2021)}]{beit}
Bao, H.; Dong, L.; and Wei, F. 2021.
\newblock {BEiT}: {BERT} Pre-Training of Image Transformers.
\newblock \emph{arXiv preprint arXiv:2106.08254}.

\bibitem[{Baradel et~al.(2022)Baradel, Br{\'e}gier, Groueix, Weinzaepfel,
  Kalantidis, and Rogez}]{posebert}
Baradel, F.; Br{\'e}gier, R.; Groueix, T.; Weinzaepfel, P.; Kalantidis, Y.; and
  Rogez, G. 2022.
\newblock PoseBERT: A Generic Transformer Module for Temporal 3D Human
  Modeling.
\newblock \emph{IEEE Transactions on Pattern Analysis and Machine
  Intelligence}.

\bibitem[{Bouazizi et~al.(2022)Bouazizi, Holzbock, Kressel, Dietmayer, and
  Belagiannis}]{MotionMixer}
Bouazizi, A.; Holzbock, A.; Kressel, U.; Dietmayer, K.; and Belagiannis, V.
  2022.
\newblock MotionMixer: MLP-based 3D Human Body Pose Forecasting.
\newblock In \emph{Proceedings of the Thirty-First International Joint
  Conference on Artificial Intelligence, {IJCAI-22}}, 791--798. International
  Joint Conferences on Artificial Intelligence Organization.

\bibitem[{B{\"u}tepage, Kjellstr{\"o}m, and Kragic(2018)}]{HRI_MOTION}
B{\"u}tepage, J.; Kjellstr{\"o}m, H.; and Kragic, D. 2018.
\newblock Anticipating many futures: Online human motion prediction and
  generation for human-robot interaction.
\newblock In \emph{International Conference on Robotics and Automation (ICRA)},
  4563--4570. IEEE.

\bibitem[{Cai et~al.(2021)Cai, Wang, Zhu, Cham, Cai, Yuan, Liu, Zheng, Yan,
  Ding, Shen, Liu, and Thalmann}]{UnifiedCVAE}
Cai, Y.; Wang, Y.; Zhu, Y.; Cham, T.-J.; Cai, J.; Yuan, J.; Liu, J.; Zheng, C.;
  Yan, S.; Ding, H.; Shen, X.; Liu, D.; and Thalmann, N.~M. 2021.
\newblock A Unified 3D Human Motion Synthesis Model via Conditional Variational
  Auto-Encoder.
\newblock In \emph{Proceedings of the IEEE/CVF International Conference on
  Computer Vision (ICCV)}, 11645--11655.

\bibitem[{Dang et~al.(2021)Dang, Nie, Long, Zhang, and Li}]{MSR-GCN}
Dang, L.; Nie, Y.; Long, C.; Zhang, Q.; and Li, G. 2021.
\newblock MSR-GCN: Multi-Scale Residual Graph Convolution Networks for Human
  Motion Prediction.
\newblock In \emph{Proceedings of the IEEE/CVF International Conference on
  Computer Vision (ICCV)}, 11467--11476.

\bibitem[{Dosovitskiy et~al.(2020)Dosovitskiy, Beyer, Kolesnikov, Weissenborn,
  Zhai, Unterthiner, Dehghani, Minderer, Heigold, Gelly et~al.}]{vit}
Dosovitskiy, A.; Beyer, L.; Kolesnikov, A.; Weissenborn, D.; Zhai, X.;
  Unterthiner, T.; Dehghani, M.; Minderer, M.; Heigold, G.; Gelly, S.; et~al.
  2020.
\newblock An image is worth 16x16 words: Transformers for image recognition at
  scale.
\newblock \emph{arXiv preprint arXiv:2010.11929}.

\bibitem[{Duan et~al.(2021)Duan, Shi, Zou, Lin, Qian, Zhang, Lab, of~Michigan,
  and NetEase}]{TRANSFORMER_MIB1}
Duan, Y.; Shi, T.; Zou, Z.; Lin, Y.; Qian, Z.; Zhang, B.; Lab, Y. Y. N. F.~A.;
  of~Michigan, U.; and NetEase. 2021.
\newblock Single-Shot Motion Completion with Transformer.
\newblock \emph{ArXiv}, abs/2103.00776.

\bibitem[{Guo et~al.(2023)Guo, Du, Shen, Lepetit, Alameda-Pineda, and
  Moreno-Noguer}]{SIMPLE}
Guo, W.; Du, Y.; Shen, X.; Lepetit, V.; Alameda-Pineda, X.; and Moreno-Noguer,
  F. 2023.
\newblock Back to MLP: A Simple Baseline for Human Motion Prediction.
\newblock In \emph{Proceedings of the IEEE/CVF Winter Conference on
  Applications of Computer Vision (WACV)}, 4809--4819.

\bibitem[{Harvey and Pal(2018)}]{RNN2_MIB}
Harvey, F.~G.; and Pal, C.~J. 2018.
\newblock Recurrent transition networks for character locomotion.
\newblock \emph{SIGGRAPH Asia 2018 Technical Briefs}.

\bibitem[{Harvey et~al.(2020{\natexlab{a}})Harvey, Yurick, Nowrouzezahrai, and
  Pal}]{lafan1dataset}
Harvey, F.~G.; Yurick, M.; Nowrouzezahrai, D.; and Pal, C. 2020{\natexlab{a}}.
\newblock Robust motion in-betweening.
\newblock \emph{ACM Transactions on Graphics (TOG)}, 39(4): 60--1.

\bibitem[{Harvey et~al.(2020{\natexlab{b}})Harvey, Yurick, Nowrouzezahrai, and
  Pal}]{LSTM_MIB_TG}
Harvey, F.~G.; Yurick, M.; Nowrouzezahrai, D.; and Pal, C.~J.
  2020{\natexlab{b}}.
\newblock Robust motion in-betweening.
\newblock \emph{{ACM} Trans. Graph.}

\bibitem[{He et~al.(2021)He, Chen, Xie, Li, Dollár, and Girshick}]{MAE}
He, K.; Chen, X.; Xie, S.; Li, Y.; Dollár, P.; and Girshick, R. 2021.
\newblock Masked Autoencoders Are Scalable Vision Learners.

\bibitem[{Hendrycks and Gimpel(2016)}]{gelu}
Hendrycks, D.; and Gimpel, K. 2016.
\newblock Gaussian error linear units (gelus).
\newblock \emph{arXiv preprint arXiv:1606.08415}.

\bibitem[{Ho, Jain, and Abbeel(2020)}]{ddpm_paper}
Ho, J.; Jain, A.; and Abbeel, P. 2020.
\newblock Denoising diffusion probabilistic models.
\newblock \emph{Advances in Neural Information Processing Systems (NeurIPS)},
  33: 6840--6851.

\bibitem[{Holden, Saito, and Komura(2016)}]{RNN1_MIB}
Holden, D.; Saito, J.; and Komura, T. 2016.
\newblock A Deep Learning Framework for Character Motion Synthesis and Editing.
\newblock \emph{ACM Trans. Graph.}

\bibitem[{Ionescu et~al.(2014)Ionescu, Papava, Olaru, and
  Sminchisescu}]{h36m_pami}
Ionescu, C.; Papava, D.; Olaru, V.; and Sminchisescu, C. 2014.
\newblock Human3.6M: Large Scale Datasets and Predictive Methods for 3D Human
  Sensing in Natural Environments.
\newblock \emph{Transactions on Pattern Analysis and Machine Intelligence
  (TPAMI)}, 36(7): 1325--1339.

\bibitem[{Jain et~al.(2016)Jain, Zamir, Savarese, and Saxena}]{sRNN}
Jain, A.; Zamir, A.~R.; Savarese, S.; and Saxena, A. 2016.
\newblock Structural-rnn: Deep learning on spatio-temporal graphs.
\newblock In \emph{Conference on Computer Vision and Pattern Recognition
  (CVPR)}, 5308--5317.

\bibitem[{Jiang, Chen, and Guo(2022)}]{dual_maskedAE}
Jiang, J.; Chen, J.; and Guo, Y. 2022.
\newblock A Dual-Masked Auto-Encoder for Robust Motion Capture with
  Spatial-Temporal Skeletal Token Completion.
\newblock In \emph{Proceedings of the 30th ACM International Conference on
  Multimedia}, MM '22, 5123–5131. New York, NY, USA: Association for
  Computing Machinery.
\newblock ISBN 9781450392037.

\bibitem[{Karamcheti et~al.(2023)Karamcheti, Nair, Chen, Kollar, Finn, Sadigh,
  and Liang}]{mae_repr_robots}
Karamcheti, S.; Nair, S.; Chen, A.~S.; Kollar, T.; Finn, C.; Sadigh, D.; and
  Liang, P. 2023.
\newblock Language-Driven Representation Learning for Robotics.

\bibitem[{Kim et~al.(2022)Kim, Byun, Shin, Won, and Choi}]{CMIB}
Kim, J.; Byun, T.; Shin, S.; Won, J.; and Choi, S. 2022.
\newblock Conditional motion in-betweening.
\newblock \emph{Pattern Recognition}, 132.

\bibitem[{Kingma, Welling et~al.(2019)}]{vae}
Kingma, D.~P.; Welling, M.; et~al. 2019.
\newblock An introduction to variational autoencoders.
\newblock \emph{Foundations and Trends in Machine Learning}, 12(4): 307--392.

\bibitem[{Kipf and Welling(2016)}]{gcn}
Kipf, T.~N.; and Welling, M. 2016.
\newblock Semi-supervised classification with graph convolutional networks.
\newblock \emph{arXiv preprint arXiv:1609.02907}.

\bibitem[{Li et~al.(2022)Li, Chang, Mishra, Zhang, Katabi, and
  Krishnan}]{mage_image}
Li, T.; Chang, H.; Mishra, S.~K.; Zhang, H.; Katabi, D.; and Krishnan, D. 2022.
\newblock Mage: Masked generative encoder to unify representation learning and
  image synthesis.
\newblock \emph{arXiv preprint arXiv:2211.09117}.

\bibitem[{Li et~al.(2023)Li, Rao, Pan, Wang, and Xu}]{ti_mae}
Li, Z.; Rao, Z.; Pan, L.; Wang, P.; and Xu, Z. 2023.
\newblock Ti-MAE: Self-Supervised Masked Time Series Autoencoders.

\bibitem[{Ma et~al.(2022)Ma, Nie, Long, Zhang, and Li}]{ST-DGCN}
Ma, T.; Nie, Y.; Long, C.; Zhang, Q.; and Li, G. 2022.
\newblock Progressively Generating Better Initial Guesses Towards Next Stages
  for High-Quality Human Motion Prediction.
\newblock In \emph{Proceedings of the IEEE/CVF Conference on Computer Vision
  and Pattern Recognition (CVPR)}, 6437--6446.

\bibitem[{Mao, Liu, and Salzmann(2020)}]{mao2020history}
Mao, W.; Liu, M.; and Salzmann, M. 2020.
\newblock History repeats itself: Human motion prediction via motion attention.
\newblock In \emph{European Conference on Computer Vision (ECCV)}, 474--489.
  Springer.

\bibitem[{Mao et~al.(2019)Mao, Liu, Salzmann, and Li}]{DCT-GCN}
Mao, W.; Liu, M.; Salzmann, M.; and Li, H. 2019.
\newblock Learning trajectory dependencies for human motion prediction.
\newblock In \emph{International Conference on Computer Vision (ICCV)},
  9489--9497.

\bibitem[{Martinez, Black, and Romero(2017)}]{RNN_Motion2}
Martinez, J.; Black, M.~J.; and Romero, J. 2017.
\newblock On human motion prediction using recurrent neural networks.
\newblock In \emph{Conference on Computer Vision and Pattern Recognition
  (CVPR)}, 2891--2900.

\bibitem[{Oreshkin et~al.(2022)Oreshkin, Valkanas, Harvey, Ménard, Bocquelet,
  and Coates}]{DELTA_INTERP}
Oreshkin, B.~N.; Valkanas, A.; Harvey, F.~G.; Ménard, L.-S.; Bocquelet, F.;
  and Coates, M.~J. 2022.
\newblock Motion Inbetweening via Deep $\Delta$-Interpolator.
\newblock arXiv:2201.06701.

\bibitem[{Tevet et~al.(2022)Tevet, Raab, Gordon, Shafir, Bermano, and
  Cohen-Or}]{MDM}
Tevet, G.; Raab, S.; Gordon, B.; Shafir, Y.; Bermano, A.~H.; and Cohen-Or, D.
  2022.
\newblock Human Motion Diffusion Model.
\newblock \emph{arXiv preprint arXiv:2209.14916}.

\bibitem[{Tong et~al.(2022)Tong, Song, Wang, and Wang}]{videomae}
Tong, Z.; Song, Y.; Wang, J.; and Wang, L. 2022.
\newblock Video{MAE}: Masked Autoencoders are Data-Efficient Learners for
  Self-Supervised Video Pre-Training.
\newblock In \emph{Advances in Neural Information Processing Systems}.

\bibitem[{Valls~Mascaro et~al.(2022)Valls~Mascaro, Ma, Ahn, and Lee}]{2CH-TR}
Valls~Mascaro, E.; Ma, S.; Ahn, H.; and Lee, D. 2022.
\newblock Robust human motion forcasting using transformer-based model.
\newblock In \emph{2022 IEEE/RSJ International Conference on Intelligent Robots
  and Systems (IROS)}, 10674--10680.

\bibitem[{Vaswani et~al.(2017)Vaswani, Shazeer, Parmar, Uszkoreit, Jones,
  Gomez, Kaiser, and Polosukhin}]{transformer}
Vaswani, A.; Shazeer, N.; Parmar, N.; Uszkoreit, J.; Jones, L.; Gomez, A.~N.;
  Kaiser, {\L}.; and Polosukhin, I. 2017.
\newblock Attention is all you need.
\newblock \emph{Advances in neural information processing systems (NeurIPS)},
  30.

\bibitem[{Zheng et~al.(2022)Zheng, Yang, Mo, Li, Yu, Liu, Liu, and
  Guibas}]{gimo}
Zheng, Y.; Yang, Y.; Mo, K.; Li, J.; Yu, T.; Liu, Y.; Liu, C.~K.; and Guibas,
  L.~J. 2022.
\newblock Gimo: Gaze-informed human motion prediction in context.
\newblock In \emph{Computer Vision--ECCV 2022: 17th European Conference, Tel
  Aviv, Israel, October 23--27, 2022, Proceedings, Part XIII}, 676--694.
  Springer.

\bibitem[{Zhou et~al.(2019)Zhou, Barnes, Lu, Yang, and Li}]{ortho6d}
Zhou, Y.; Barnes, C.; Lu, J.; Yang, J.; and Li, H. 2019.
\newblock On the Continuity of Rotation Representations in Neural Networks.
\newblock In \emph{Proceedings of the IEEE/CVF Conference on Computer Vision
  and Pattern Recognition (CVPR)}.

\bibitem[{Zhu et~al.(2022)Zhu, Ma, Liu, Liu, Wu, and Wang}]{motionbert}
Zhu, W.; Ma, X.; Liu, Z.; Liu, L.; Wu, W.; and Wang, Y. 2022.
\newblock Learning Human Motion Representations: A Unified Perspective.
\newblock \emph{arXiv preprint arXiv:2210.06551}.

\end{thebibliography}

\end{document}